\documentclass[runningheads]{llncs}
\usepackage{graphicx}

\usepackage{latexsym,amsmath}
\usepackage{algorithm}
\usepackage{algpseudocode}
\let\-=\mathbf
\let\@=\mathcal
\begin{document}
\title{Bayesian Optimization with Local Search\thanks{This work was partially
supported by the NSFC under grant number 11301337.}}
%
%
\author{Yuzhou Gao\inst{1} \and
Tengchao Yu\inst{1} \and
Jinglai Li\inst{2}}
\authorrunning{Y. Gao et al.}
%
\institute{Shanghai Jiao Tong University, Shanghai 200240, China \\
 \and
University of Birmingham, Edgbaston
Birmingham, B15 2TT, UK\\
\email{j.li.10@bham.ac.uk}}
\maketitle              
\begin{abstract}
  Global optimization finds applications in a wide range of real world problems. 
The multi-start methods are a popular class of global optimization techniques, which are based on
the idea of conducting local searches at multiple starting points.
In this work we propose a new multi-start algorithm where the starting points are determined
in a Bayesian optimization framework. 
Specifically, the method can be understood as to construct a new function by conducting
local searches of the original objective function, where the new function attains the same global optima
as the original one. Bayesian optimization is then applied to find the global optima of the 
new local search defined function.

\end{abstract}
\section{Introduction}
Global optimization (GO) is a subject of tremendous potential applications, and has been
an active research topic since.
There are several difficulties associated with solving a global optimization problem:
the objective function may be expensive to evaluate and/or subject to random noise, 
it may be a black-box model and the gradient information is not available,
and the problem may admit a very large number of local minima, etc. 
In this work we focus on the last issue: namely, in many practical global optimization problems, 
it is often possible to find a local minimum efficiently, especially when the gradient information of the objective function is available, while the main challenge is to escape from a local minimum and find the global solution. 
Many metaheuristic GO methods, such simulated annealing~\cite{kirkpatrick1983optimization} 
and genetic algorithm~\cite{holland1992adaptation}, can avoid being trapped by a local minimal, but these methods do not take advantage of the property that a local problem can be quite efficiently solved, which makes them less 
efficient in the type of problems mentioned above. 

A more effective strategy for solving such problems is to combine global and local searches, 
and the multi-start (MS) algorithms~\cite{marti2016multi} have become a very popular class of methods along this line.  
Loosely speaking the MS algorithms attempt to find a global solution by performing local optimization from multiple starting points. 
Compared to search based global optimization algorithms, the (MS) methods is particularly suitable for problems where a local optimization can be performed efficiently. 
The most popular MS methods include the clustering~\cite{kan1987stochastic,tu2002studies} and the Multi Level Single Linkage (MLSL)~\cite{rinnooy1987stochastic} methods
and the OptQuest/NLP algorithm~\cite{ugray2007scatter}.
More recently, new MS algorithms have been proposed and applied to machine learning problems~\cite{gyorgy2011efficient,kawaguchi2016global}. 
One of the  most important issues in a MS algorithm is how to determine the initial points, i.e. the points to start a local search (LS) from. Most MS algorithms determine the initial points sequentially, which in each step 
requires to find the next initial point based on the current information. We shall adopt this setup in this work 
and so the question we want to address in the present work is 
\textit{how to determine the next ``best'' initial point given the information at the current step.} 

The main idea presented in this work is
to sequentially determine the starting points in a Bayesian optimization (BO)~\cite{shahriari2015taking,snoek2012practical,mockus2012bayesian} 
framework.
The standard BO algorithm is designed to solve a global optimization problem directly
without using LS:
it uses a Bayesian framework and an experimental design strategy to search for 
the global minimizers.
The BO algorithms have found success in many practical GO problems,  
especially  for those expensive and noisy objective functions~\cite{brochu2010tutorial}. 
Nevertheless, the BO methods do not take advantage of efficient local solvers even when that is possible. 
 In this work, instead of applying BO directly to the global optimization problem, we propose to use it to identify
 starting points  for the local solvers in a MS formulation. 
Within the BO framework,  we can determine the starting points using a rigorous and effective
 experimental design approach.  
 
 An alternative view of the proposed method is that 
we define a new function by solving a local optimization problem of the original objective function. 
By design the newly defined function is discrete-valued and has the same global optimizers as 
the original objective function. And we then perform BO to find the global minima for the new function.
From this perspective, the method can be understood as to pair the BO method with a local solver,
and we reinstate that the method requires that the local problems can be solved efficiently.     
For example, in many statistical learning problems with large amounts of data, 
a noisy estimate of the gradients can be computed more efficiently than the evaluation of the 
objective function~\cite{bottou2010large}, 
and it follows that a local solution can be obtained at a reasonable computational cost.    


The rest of the work is organized as follows. 
In Section~\ref{sec:method} we introduce the MS algorithms for GO problems, 
and present our BO based method to identify the starting points. 
In Section \ref{sec:examples} we provide several examples to demonstrate the performance of the proposed method.
Finally Section \ref{sec:conclusions} offers some closing remarks.

\section{Bayesian optimization with local search}\label{sec:method}

\subsection{Generic multi-start algorithms}
Suppose that we want to solve a bound constrained optimization problem:
\begin{equation}
\min_{\-x \in \Omega} f(\-x),
\end{equation}
where $\Omega$ is a compact subspace of $R^n$.
In general, the problem may admit multiple local minimizers and we want to find the global solution of it. 
As has been mentioned earlier, the MS algorithms are a class of GO methods for problems where
LS can be conducted efficiently. 
The MS iteration consists of two steps: a global step where an initial point is generated,
and a local step which performs a local search from the generated initial point.
A pseudocode of the generic MS algorithm is given in Alg.~\ref{alg:ms}. 
It can be seen here that one of the key issues of the MS algorithm is how to generate the starting point in each iteration.
 A variety of methods have been proposed to choose the starting points,
 and they are usually designed for different type of problems. 
For example, certain methods such as \cite{ugray2007scatter} assume that the evaluation of the objective function
 is much less computationally expensive than the local searches, 
 and as a result they try to reduce the number of local searches at the price of conducting a rather large number of function evaluations in the state space.  
On the other hand, in another class of problems, 
a satisfactory local solution may be obtained at a reasonable computational cost, 
and as will be discussed later we shall use the BO algorithm to determine the initial points.
For this purpose, we next give a brief overview of BO.

\begin{algorithm}
    \caption{A generic MS algorithm}
    \label{alg:ms}
    \begin{algorithmic}[1]
	\let\-=\mathbf

\State set $i=0$;
 \While {Stopping criteria are not satisfied} 
\State $n=n+1$;
\State generate a new initial point $\-x_n$  based on some prescribed rules;
\State perform a LS from $\-x_n$ and store the obtained local minimal value;
\EndWhile
\State\Return the smallest local minimum value found.

   \end{algorithmic}
\end{algorithm}
\subsection{Bayesian Optimization}
The Bayesian optimization (BO) is very popular global optimization method, which treats 
the objective function as a blackbox. 
Simply put, BO involves the use of a probabilistic model that defines a distribution over objective function.
In practice the probabilistic model is usually constructed with the Gaussian Process (GP) regression:
namely the function $f(\-x)$ is assumed to be a Gaussian process defined on $\Omega$, 
the objective function is queried at certain locations, 
and the distribution of the function value at any location $\-x$, conditional on the observations,
which is Gaussian, can be explicitly computed from the Bayesian formula. 
Please see Appendix~\ref{sec:gp} for a brief description of the GP construction. 
Based on the current GP model of $f(\-x)$ the next point to query is determined in an experimental design formulation. 
Usually
the point to query is determined by maximizing an acquisition function $\alpha(\-x, \hat{f})$ where
$\hat{f}$ is the GP model of $f$, which is designed based on the exploration and the exploitation purposes of the algorithm.  
Commonly used acquisition functions include the Expected Improvement, the Probability 
of Improvement, and the Upper Confidence Bound,
and 
interested readers may consult \cite{snoek2012practical} for detailed discussions and comparisons of these acquisition functions. We describe the standard version of BO in Alg.~\ref{alg:bo}.
	
	\begin{algorithm}
    \caption{The BO algorithm}
    \label{alg:bo}
    \begin{algorithmic}[1]
		\let\-=\mathbf
\State generate a number of points $\{x_1,...,x_{N_0}\}$ in $\Omega$.
\State evaluate $y_n=f(\-x_n)$ for $n=1:N_0$;
\State let $D_{N_0}=\{(\-x_n,y_n)\}_{n=1}^{N_0}$;
\State construct a GP model from $D_{N_0}$, denoted as $\hat{f}_{N_0}$;
\State $n=N_0$;
 \While {stopping criteria are not satisfied} 
\State   $\-x_{n+1} = \arg \max \alpha(\-x; \hat{f}_n)$
\State  $y_{n+1}=f(\-x_{n+1})$;
\State augment data $D_{n+1} = D_n\cup\{(\-x_{n+1}, y_{n+1})\}$
\State update GP model obtaining $\hat{f}_{n+1}$;
\State $n=n+1$;
\EndWhile
\State\Return $y_{\min} =\min \{y_n\}_{n=1}^N$;

   \end{algorithmic}
\end{algorithm}

\subsection{The BO with LS algorithm}
\begin{figure}
\centerline{\includegraphics[width=0.65\textwidth]{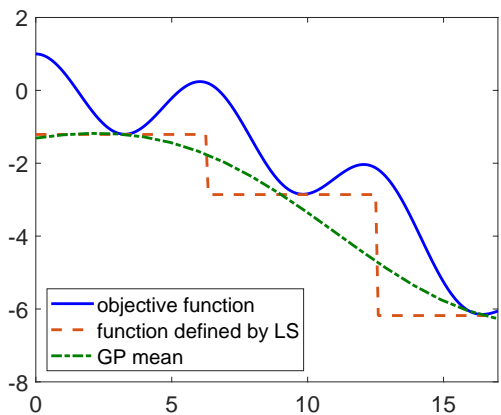}}
\caption{A schematic illustration of the BOwLS algorithm: the solid line
is the original objective function, the dashed line 
is the function defined by LS, and the dashed-dotted line is the GP regression 
of the LS defined function.} \label{f:bowls}
\end{figure}

Now we present our method that integrate MS and BO.
The idea behind the method is rather simple: we perform BO for a new function which has the same global minima as the original function $f(\-x)$.
The new function is defined via conducting local search of $f(\-x)$.
Specifically suppose we have local solver $\mathcal{L}$ defined as,
\begin{equation}
\-x^* = \mathcal{L}(f(\cdot),\-x),\label{e:ls}
 \end{equation}
where $f(\cdot)$ is the objective function, $\-x$ is the initial point of the local search,
and $\-x^*$ is the obtained local minimal point. 
$\mathcal{L}$ can represent any local optimization approach, with or without gradient,
and we require that for any given initial point $\-x^*$, the solver $\@L$ will 
return a unique local minimum $\-x^*$. 
Using both $\@L$ and $F$, we can define a new function
\begin{equation}
y=F_{\@L}(\-x) = f(\-x^*),
\end{equation}
where $\-x^*$ is the output of Eq.~\eqref{e:ls} 
with objective function $f$ and initial point $\-x$. 
That is, the new function $F_{\@L}$ takes a starting point $\-x$ 
as its input, and returns the local minimal value of $f$ found by the local solver $\@L$ as its output. 
It should be clear that $F_{\@L}$ is a well-defined function on $R^n$, which has 
the same global minima as function $f(\-x)$. 
Moreover, suppose that $f(\-x)$ only has a finite number of local minima, and $F_{\@L}(\-x)$ is
discrete-valued. Please see Fig.~\ref{f:bowls} for a schematic illustration of the 
new function defined by LS and its GP approximation. 
Next we apply standard BO algorithm to the newly constructed function $F_\@L(\-x)$, 
and the global solution of $F_{\@L}$ found by BO is regarded as the global solution of $f(\-x)$. 
We refer to the proposed algorithm as BO with LS (BOwLS) 
and we provide the complete procedure of it in Alg.~\ref{alg:bowls}.
We reinstate that, as one can see from the algorithm,  BOwLS is essentially a MS scheme, which uses the BO experimental design criterion 
to determine the next starting point. 
When desired, multiple starting points can also be determined in the BO framework, and we refer to the aforementioned 
BO references for details of this matter. 

\begin{algorithm}
    \caption{The BOwLS algorithm}
    \label{alg:bowls}
		\let\-=\mathbf
    \begin{algorithmic}[1]

\State let $D_0=\emptyset$;
 \For {$n=1$:$N_0$}
\State solve $[y^*,\-x^*]=\@L(f(\-x),\-x_{n})$;
\State let $y_{n}=y^*$;
\State augment data $Dn = D_{n-1}\cup\{(\-x_{n}, y_{n})\}$
\EndFor

\State construct a GP model from $D_{N_0}$, denoted as $\hat{f}_{N_0}$;
\State $n=N_0$;
 \While {stopping criteria are not satisfied} 
\State   $\-x_{n+1} = \arg \max \alpha(\-x; \hat{f}_n)$
\State solve $[y^*,\-x^*]=\@L(f(\-x),\-x_{n+1})$;
\State let $y_{n+1}=y^*$;
\State augment data $D_{n+1} = D_n\cup\{(\-x_{n+1}, y_{n+1})\}$;
\State update GP model obtaining $\hat{f}_{n+1}$;
\State $n=n+1$;
\EndWhile
\State\Return $y_{\min} =\min \{y_i\}_{i=1}^N$
   \end{algorithmic}
\end{algorithm}

\section{Numerical examples}\label{sec:examples}
In this section, we provide several mathematical and practical examples to demonstrate the performance of the proposed method. 
In each example, we solve the GO problem with three methods: MLSL, the efficient multi-start (EMS) in \cite{gyorgy2011efficient}
and the BOwLS method proposed in this work.

\subsection{Mathematical test functions} \label{sec:testfun}
\begin{figure}
\centerline{\includegraphics[width=0.75\textwidth]{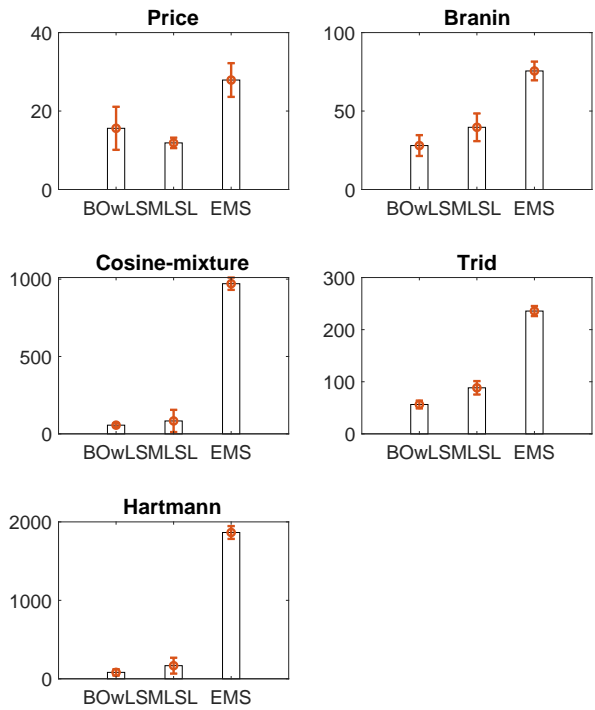}}
\caption{The average numbers of the combined evaluations 
to reach the global optima for all the three methods (the error bars indicate the standard deviations)
in the first five functions. } \label{f:testfun}
\end{figure}

\begin{figure}
\centerline{\includegraphics[width=0.5\textwidth]{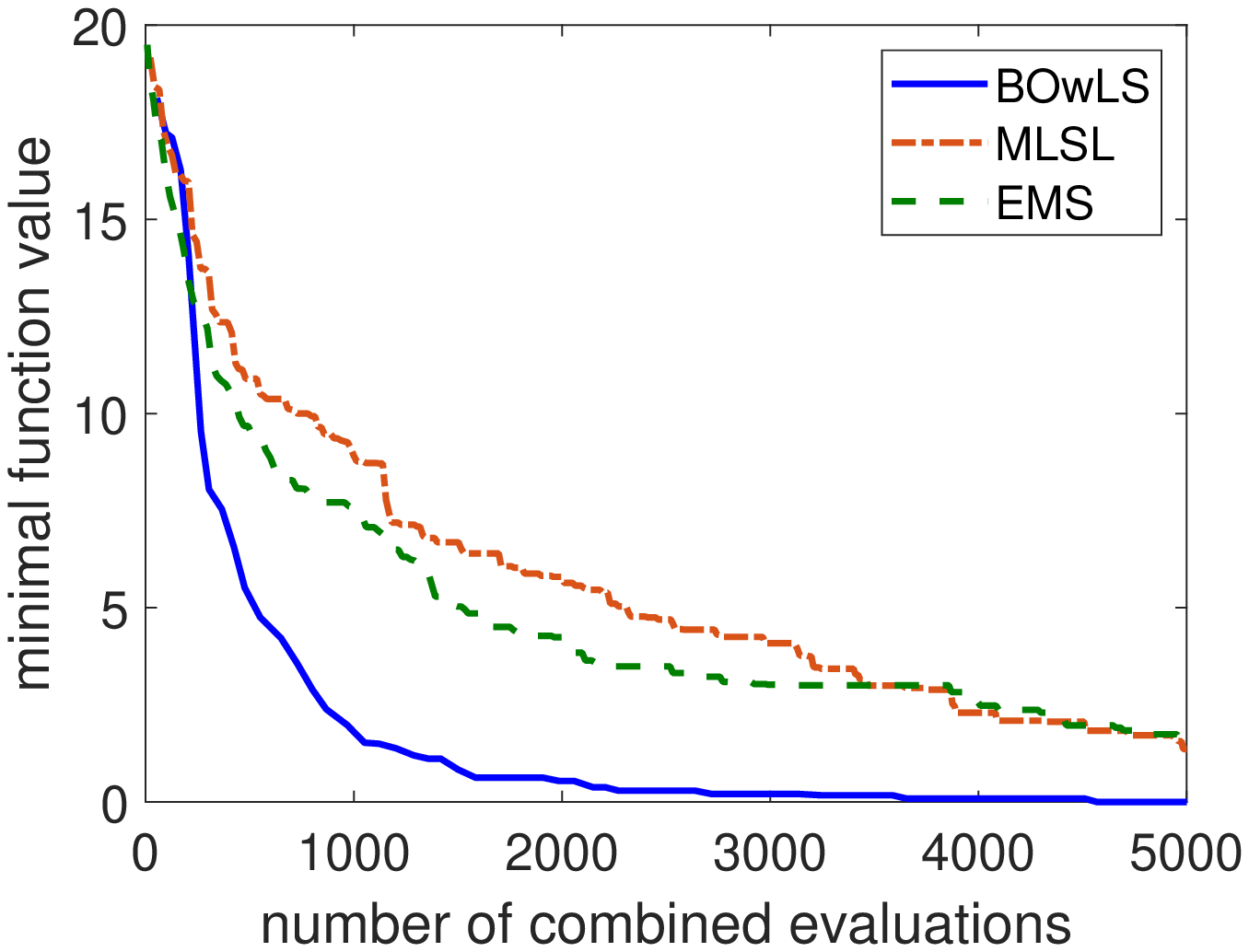}
\includegraphics[width=0.5\textwidth]{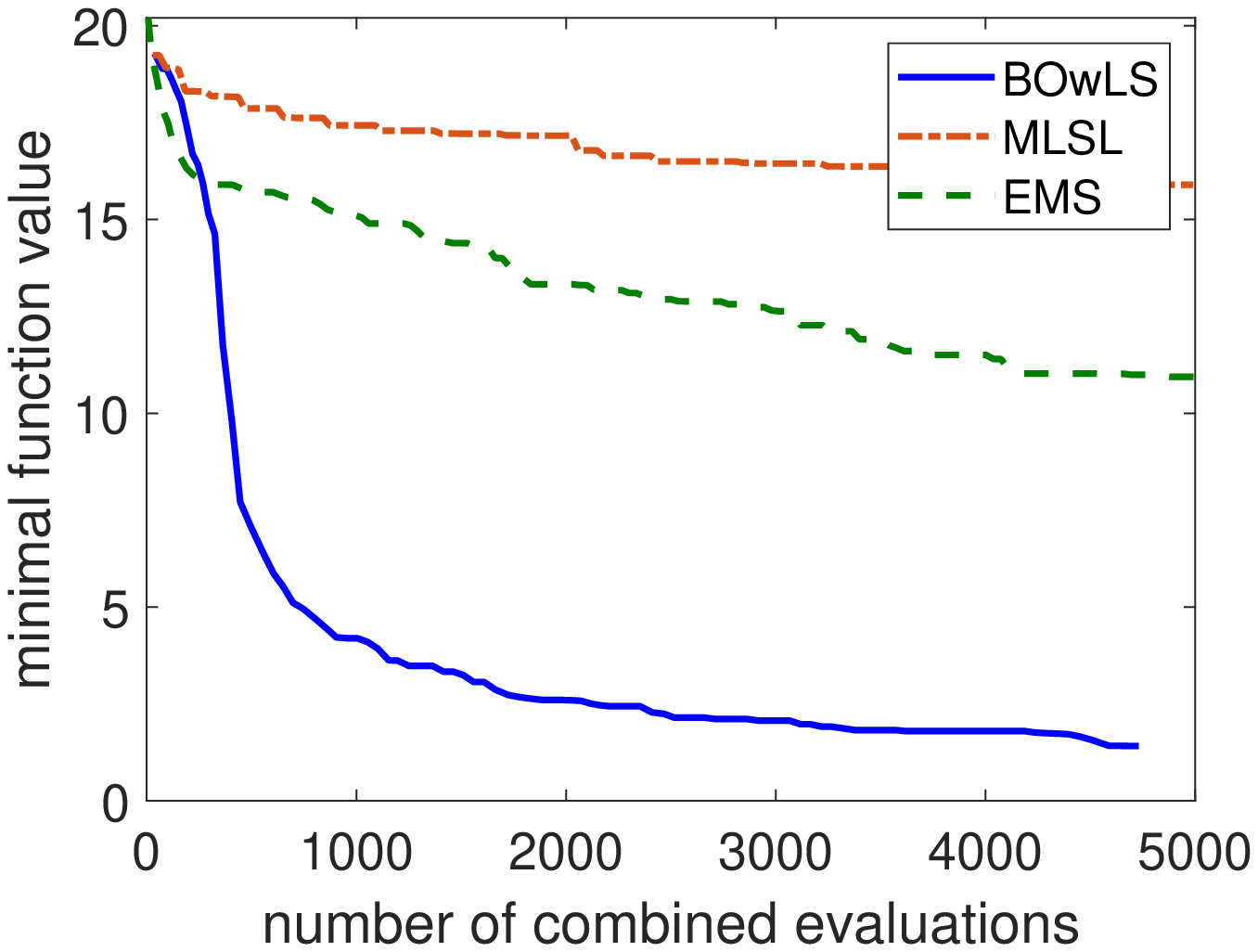}}
\caption{The minimal function value 
plotted agains the number of function/gradient combined evaluations,
for the 2-D and 4-D Ackley example. } \label{f:ackley2d}
\end{figure}

We first consider six mathematical examples that are commonly used as 
the benchmarks for GO algorithms, selected from \cite{gavana2016global}. 
 The objective functions, the domains and the global optimal solutions of these functions 
are provided in Appendix~\ref{sec:mathfun}. 
As is mentioned earlier, we solve these problems with MLSL, EMS and BOwLS methods,
and, since all these algorithms are subject to certain randomness,
we repeat the numerical experiments for 50 times.  
The  local search is conducted with the conjugate gradient method using the SciPy package~\cite{Jones:2001aa}.
In these examples we shall assume that  evaluating the objective function 
and its gradient is of similar computational cost, and so we measure the total computational 
cost  by summation of the number of function evaluations and that of the gradient evaluations. 
For test purpose, we set the stopping criterion to be that 
the number of function/gradient combined evaluations exceeds 10,000.
In our tests, we have found that all the three methods can reach 
the actual global optima within the stopping criterion in the first five functions. 
In Figs.~\ref{f:testfun} we compare the average numbers of the combined evaluations 
(and their standard deviations)
to reach the global optimal value for all the three methods in the first five test functions.  
As we can see that in all these five test functions except the example~(Price), the proposed BOwLS algorithm 
requires the least computational cost to research the global minima.  We also note that it seems that EMS requires
significantly more combined evaluations than the other two methods,
and we believe that the reason is that EMS is particularly designed for problems with a 
very large number of local minima, and these test functions are not in that case. 
On the other hand, the last example~(Ackley) is considerably more complicated, and so 
in our numerical tests all three methods have trials that can not reach the global minimum within the prescribed cost limit. 
To compare the performance of the methods, we plot the minimal function value obtained against
the number of function/gradient combined evaluations
in Fig.~\ref{f:ackley2d}. 
First the plots show that, in this example, the EMS method performs better than MLSL in both 2-D and 4-D cases,
due to the fact that this function is subject to more local minima. 
More importantly, as one can see, in both cases BOwLS performs considerably better than both EMS and MLSL,
and the advantage is more substantial in the 4-D case, suggesting that BOwLS may become more useful for complex objective functions. 

\subsection{Logistic regression}
Finally consider a Logistic regression example. 
Logistic regression is a common tool for binary regression (or classification). Specifically suppose that
we have binary regression problem where the 
output takes values at $y=0$ or $y=1$, 
and the probability that $y=1$ is assumed to be the the form of,  
    \begin{equation}
        h_{\-w}(\-x) = \frac{1}{1+\exp(- \sum_{i=1}^m{x_i w_i} -w_0)},
    \end{equation}
    where $\-x=(x_1,...,x_m)$ are the predictors and $\-w=(w_0,...w_m)$ are the coefficients to be determined from data. 
    The cost function for the logistic regression is taken to be 
    \[
C(h_{\-w}(\-x),y) =
\begin{cases}
-\log(h_{\-w}(\-x)) & \text{if y = 1} \\
-\log(1-h_{\-w}(\-x)) & \text{if y = 0}
\end{cases}.
\]
Suppose that we have a training set $\{(\-x_i,y_i)\}_{i=1}^n$,
and we then determine the parameters $\-w$ by solving the following optimization problem,
\begin{equation}
\min_{\-w\in W} \sum_{i=1}^n C(h_{\-w}(\-x_i),y_i).\label{e:emr}
\end{equation}
where $W$ is the domain of $\-w$. 

\begin{figure}
\leftline{\includegraphics[width=0.5\textwidth]{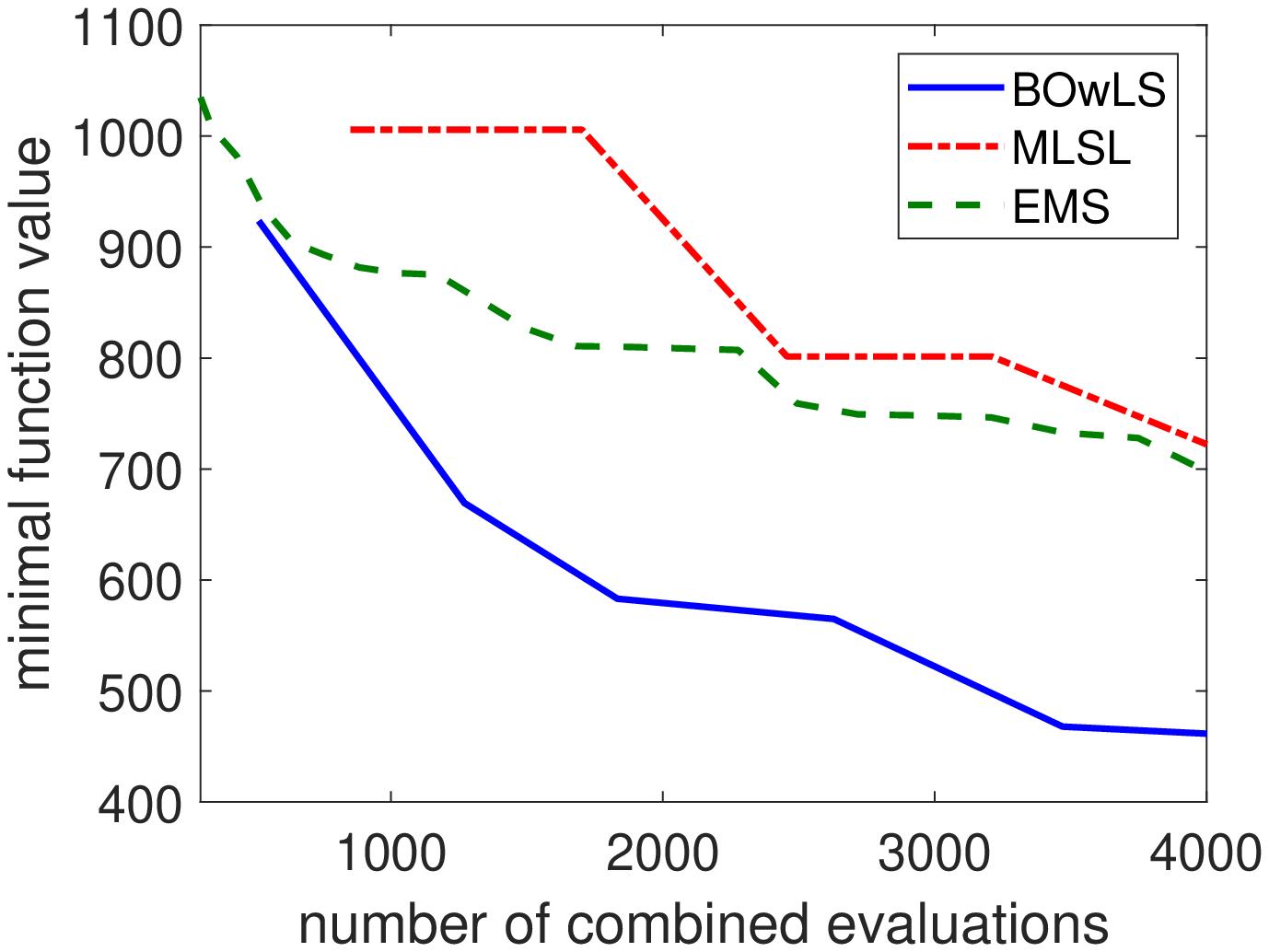}\includegraphics[width=0.5\textwidth]{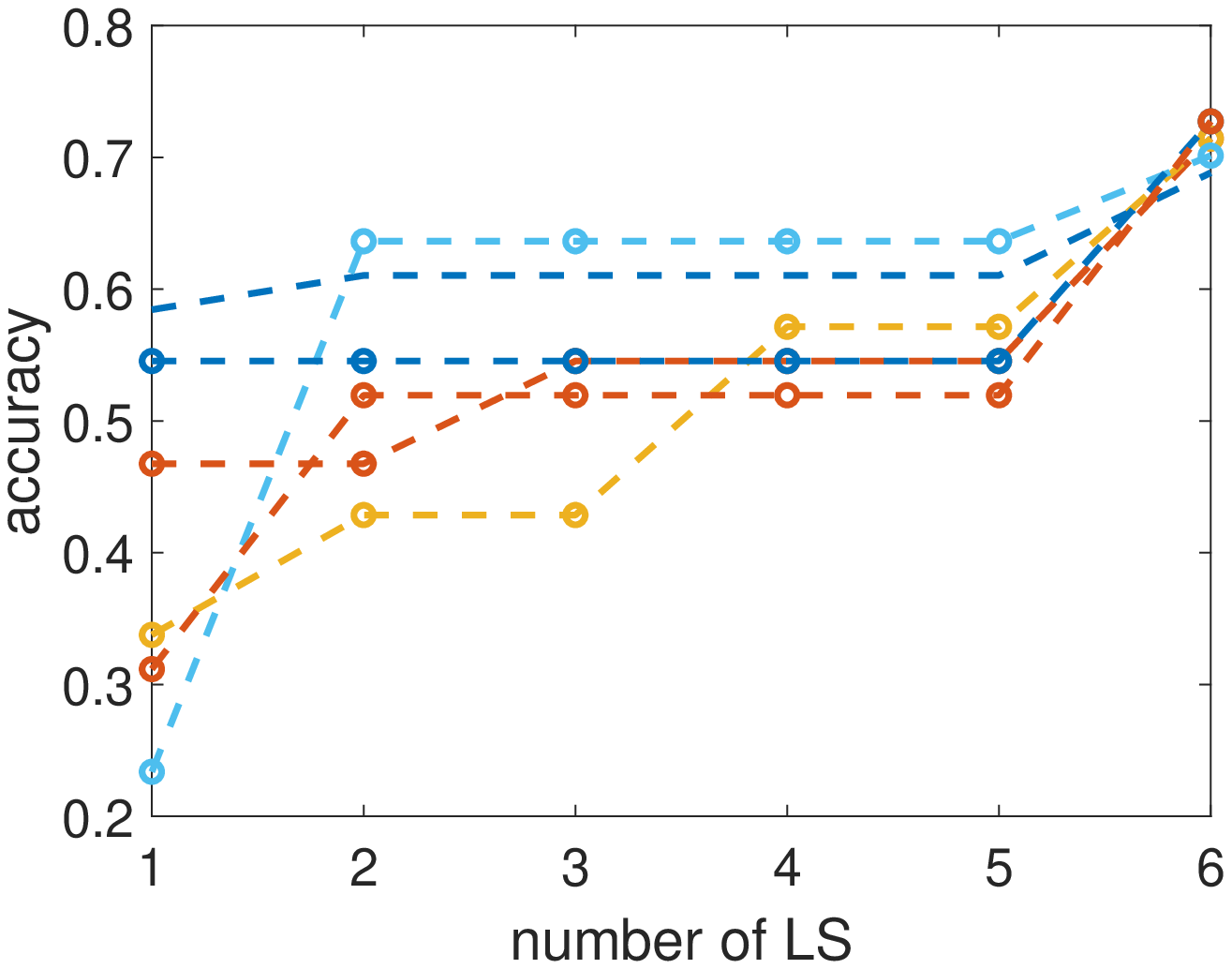}}
\caption{The minimal function value 
plotted against the number of function/gradient combined evaluations,
for the Logistic regression example. } \label{f:lr1}
\end{figure}

In this example we apply the Logistic regression to the Pima Indians Diabetes dataset \cite{nr}, the goal of which is to diagnose whether a patient has diabetes based on  8 diagnostic measures provided in the data set. The data set contains 768 instances and we split it into a  training set of 691 instances and
a test set of $77$ ones. We solve the result optimization problem~\eqref{e:emr} with the three GO algorithms,
and we repeat the computations for 100 times as before. 
The minimal function value averaged over the 100 trials is plotted against the number of combined evaluations 
in Fig.~\ref{f:lr1} (left). In this example, the EMS method actual performs better than MLSL, 
while BOwLS has the best performance measured by the number of the function/gradient combined evaluations. 
Moreover, in the BOwLS method, we expect that as the iteration approaches to the 
global optimum, the resulting Logistic model should become better and better. 
To show this, we plot in Fig.~\ref{f:lr1} (right) the prediction accuracy of the resulting model as a function 
of the BO iterations (which is also the number of LS), in six randomly selected trials out of 100. 
The figure shows that the prediction accuracy varies (overally increases) as the number of LS increases, 
which is a good evidence that the objective function in this example admits multiple local optima  
and the global optimum is needed for the optimal prediction accuracy.  

\section{Conclusions}\label{sec:conclusions}
In summary, we have presented a MS algorithm where
the starting points of local searches are
determined  by a BO framework. 
A main advantage of the method is that the BO framework
allows one to sequentially determine the next starting points
in a rigorous and effective experimental design formation.
With several numerical examples, we demonstrate that the proposed BOwLS method
has highly competitive performance against many commonly used MS algorithms. 
A major limitation of BOwLS is that, as it is based
on the BO framework, it may have difficulty in dealing with very high dimensional problems.  
We note however that a number of dimension reduction based approaches~\cite{djolonga2013high,kandasamy2015high} 
have been proposed to enable BO for high dimensional problems,
and we hope that these approaches can be extended to BOwLS as well.  
In addition, another problem that we plan to work on 
in the future to combine the BOwLS framework with 
 the stochastic gradient descent type of algorithms
to develop efficient GO algorithms for statistical learning problems.    

\appendix
\section{Construction of the GP model}\label{sec:gp}

Given the data ste $D=\{(\-x_j,y_j)\}_{j=1}^n$, the GP regression performs a nonparametric regression in a Bayesian framework~\cite{williams2006gaussian}.  
The main idea of the GP method is to assume that the data points  and the new point $(\-x,y)$ are
from a Gaussian Process defined on $R^{n_x}$,
whose mean is $\mu(\-x)$ and covariance kernel is $k(\-x,\-x')$.
Under the GP model, one can obtain directly the conditional distribution $\pi(y|\-x,D)$ that is Gaussian: 
%
  $\pi(y|\-x) =\mathcal{N}(\mu_\mathrm{GP}, \sigma^2_\mathrm{GP})$,
where the posterior mean and variance are, 
\begin{align*}
&\mu_\mathrm{GP}(\-x)=\mu(\-x)+k(\-x,\-X)(k(\-X,\-X)+\sigma_n^2I)^{-1}(\-y-\mu(\-x)),\\
&\sigma^2_\mathrm{GP} = k(\-x,\-x)-k(\-x,\-X)(k(\-X,\-X)+\sigma_n^2I)^{-1}k(\-X,\-x). 
\end{align*}
Here $\-y^* = \left[ {y}_1, \ldots, {y}_n\right]$, $\-X = \left[\-x_1, \ldots, \-x_n\right]$, $\sigma_n^2$ is the variance of observation noise, $I$ is 
an identity matrix, 
and the notation $k(\-A,\-B)$ denotes the matrix of the covariance evaluated at all pairs of points in set $\-A$ and in set $\-B$ using 
the kernel function $k(\cdot,\cdot)$.
In particular, if the data points are generated according to an underlying function $f(x)$ (which is the objective function in the BO setting), the distribution $\pi(y|\-x)$ then provides a probabilistic characterization of the funtion $f(\-x)$ 
which can be used to predict the function value of $f(\-x)$ as well as quantify the uncertainty in the prediction. 
In Section~\ref{sec:method}, we refer to this probabilistic characterization, i.e., the Gaussian distribution $\pi(y|\-x)$ 
as $\hat{f}$. 
There are a lot of technical issues of the GP construction, such as how to choose the kernel functions
and determine the hyperparameters,  
are left out of this paper, 
and for more details of the method, we refer the readers to~\cite{williams2006gaussian}. 

\section{The mathematical test functions}
The test functions used in Section~\ref{sec:testfun} are: 
\medskip

 \noindent Price (2-D):
    \[
    f(x) = 1 + \sin^2(x_1) + \sin^2(x_2) - 0.1e^{-x_1^2-x_2^2}.
    \]
    Branin (1-D):
    \[
        f(x) = (-1.275\frac{x_1^2}{\pi^2} + 5\frac{x_1}{\pi} + x_2 -6)^2 + (10-\frac{5}{4\pi})\cos(x_1) + 10.
    \]
    Cosine-mixture (4-D):
    \[
        f(\-x) = -0.1\sum_{i=1}^4 \cos(5\pi x_i) - \sum_{i = 1}^4 x_i^2.
    \]
    Trid (6-D):
    \[
        f_{Trid}(\-x) = \sum_{i=1}^6 (x_i - 1)^2 - \sum_{i=2}^6 x_i x_{i-1}.
    \]
    Hartmann (6-D):
    \[
        f(\mathbf{x}) = -\sum_{i=1}^{4} c_i \exp{(-\sum_{j=1}^{6}a_{ij}(x_j - p_{ij})^2)},
    \]
where
    \begin{align*}
    &a= 
    \begin{pmatrix}
    10.0  & 3.0  & 17.0 & 3.50  & 1.70  & 8.0   \\0.05  & 10.0 & 17.0 & 0.10  & 8.00  & 14.0  \\ 3.0  & 3.50 & 1.70 & 10.0  & 17.0 & 8.0   \\17.0 & 8.0 & 0.05 & 10.0 & 0.10  & 14.0 \\
    \end{pmatrix},\,
    c = 
    \begin{pmatrix}
    1.0\\1.2\\3.0\\3.2\\
    \end{pmatrix},\\
   &p = 
    \begin{pmatrix}
    0.1312 & 0.1696 & 0.5569 & 0.0124 & 0.8283 & 0.5886 \\0.2329 & 0.4135 & 0.8307 & 0.3736 & 0.1004 & 0.9991 \\0.2348 & 0.1451 & 0.3522 & 0.2883 & 0.3047 & 0.6650 \\0.4047 & 0.8828 & 0.8732 & 0.5743 & 0.1091 & 0.0381 \\
    \end{pmatrix}.
\end{align*}
    Ackley ($n$-D):
    \[
    f(x) = -20 e^{-0.2\sqrt{\frac{1}{n} \sum_{i=1}^n x_i^2}} - e^{\frac{1}{n} \sum_{i=1}^n \cos(2\pi x_i)} + 20 + e,
    \]
where $n$ is taken to be $2$ and $4$ respectively. 
       
The domains and global minimal values of these  functions are shown in Table~\ref{tb:testfun}.
    \begin{table}
        \centering
        \begin{tabular}{c|c|c}
        \hline
        Functions & domain &minimal value\\ \hline
             Price& $[-10,10]^2$ & -3\\ \hline
             Branin &$ [-5,10] \times [0,15]$&0.397 \\ \hline
             Cosine-mixture 4d &$ [-1,1]^4$ & -0.252 \\ \hline
             Trid  & $[-20,20]^6$ & -50 \\ \hline
             Hartmann & $[0,1]^6$ & -3.323 \\ \hline
             Ackley & $[-32.768,32.768]^n$ & 0 \\ \hline
        \end{tabular}
        \caption{The domains and the global minimal values of the test functions.}
        \label{tb:testfun}
    \end{table}

\label{sec:mathfun}

\section*{Acknowledgement} This work was partially
supported by the NSFC under grant number 11301337.

 \bibliographystyle{splncs04}
 \bibliography{bowls}
\end{document}